\documentclass[letterpaper, 10 pt, conference]{ieeeconf}

\IEEEoverridecommandlockouts                              

\usepackage{amsmath} 
\usepackage{url}
\usepackage{booktabs}
\usepackage{hyperref}
\usepackage{cleveref}
\usepackage{graphicx}
\usepackage{tabularx}
\usepackage{tablefootnote}
\usepackage{pifont}
\usepackage{xcolor}
\usepackage{pgf-pie}

\title{\LARGE \bf
TR2MTL: LLM based framework for Metric Temporal Logic Formalization of Traffic Rules
}

\author{Kumar Manas, Stefan Zwicklbauer and Adrian Paschke
\thanks{K. Manas is with the Freie Universit{\"a}t Berlin, Germany and Continental Automotive Technologies GmbH, Germany. {\tt\small kumar.manas@fu-berlin.de}.\newline
S. Zwicklbauer is with the Continental Automotive Technologies GmbH, Germany.\newline
A. Paschke is with the Freie Universit{\"a}t Berlin, Germany and Fraunhofer FOKUS Berlin, Germany.\newline
         This work is partially funded by the German Federal Ministry for Economic Affairs and Energy within the project ``KI Wissen".}
}

\usepackage{textcomp}
\usepackage[absolute,showboxes]{textpos}
\TPMargin{5pt}

\newcommand{\copyrightstatement}{
    \begin{textblock}{0.82}(0.09,0.93)
         \noindent{\footnotesize{\copyright 2024 IEEE.
         Personal use of this material is permitted.
         Permission from IEEE must be obtained for all other uses, in any current or future media, including reprinting/republishing this material for advertising or promotional purposes, creating new collective works, for resale or redistribution to servers or lists, or reuse of any copyrighted component of this work in other works.

         \vspace{2mm}
         \noindent
         Accepted for publication in Proceedings of the IEEE Intelligent Vehicles Symposium (IV), Jeju Island - Korea, 2-5 June 2024.}}
    \end{textblock}
}   

\begin{document}

\maketitle
\copyrightstatement
\thispagestyle{empty}
\pagestyle{empty}

\begin{abstract}
Traffic rules formalization is crucial for verifying the compliance and safety of autonomous vehicles (AVs). However, manual translation of natural language traffic rules as formal specification requires domain knowledge and logic expertise, which limits its adaptation. This paper introduces TR2MTL, a framework that employs large language models (LLMs) to automatically translate traffic rules (TR) into metric temporal logic (MTL). It is envisioned as a human-in-loop system for AV rule formalization. It utilizes a chain-of-thought in-context learning approach to guide the LLM in step-by-step translation and generating valid and grammatically correct MTL formulas. It can be extended to various forms of temporal logic and rules. We evaluated the framework on a challenging dataset of traffic rules we created from various sources and compared it against LLMs using different in-context learning methods. Results show that TR2MTL is domain-agnostic, achieving high accuracy and generalization capability even with a small dataset. Moreover, the method effectively predicts formulas with varying degrees of logical and semantic structure in unstructured traffic rules.

\end{abstract}

\section{INTRODUCTION}
\label{into}
Safety and liability concerns make traffic rule compliance important for the automated vehicles (AV) industry. Countries-specific traffic rulebooks such as German Road Traffic Regulation (StVO)\footnote{\url{https://www.gesetze-im-internet.de/stvo_2013/}} and general driving guides such as Vienna Convention on Road Traffic (VCoRT)\footnote{\url{https://unece.org/DAM/trans/conventn/crt1968e.pdf}} govern the rights and obligation of road users. However, traffic rules are often written in natural language, which can be vague, inconsistent, or context-dependent. This makes it hard for AVs to interpret and follow traffic rules in different situations and locations. For example, the meaning of \textit{safe distance} while driving can vary depending on the driver's preference.

A possible solution is to use formal logic, such as metric temporal logic (MTL), to represent traffic rules precisely and unambiguously. MTL can capture the temporal aspects of traffic rules~\cite{thati_monitoring_2005} and enable various applications, such as verification, planning, monitoring, or prediction. MTL is a propositional bounded-operator logic that extends propositional logic with the temporal operator. This makes it especially suitable for safety and verification tasks~\cite{alur_logics_1992} due to its ability to consider time constraints on quantitative temporal operators, unlike linear temporal logic. In the planning module, formalized traffic rules can restrict AV maneuvers as per rules and contribute to rule-compliant behavior. MTL formalization of rules is not straightforward. It requires domain knowledge and expertise in both natural language and formal logic. It also involves complex reasoning and translation processes that could be more intuitive for humans. Moreover, it takes time and effort to manually formalize rules and specifications from different sources and domains. Therefore, an automated translator that translates natural language traffic rules into MTL specifications is desirable.

Legal traffic rules to MTL can be considered a translation or semantic parsing task. We transform the semantic parsing as a ``translation" problem. However, training such a semantic parser is difficult due to the large annotated dataset requirement of natural language traffic rules and temporal logic pairs, which is expensive to obtain due to the involvement of logic experts. Unlike the conventional ``in-context learning"~\cite{gpt3} approach which relies on prompting and language understanding of large language models (LLMs) to minimize the need for large datasets, our method addresses the limitations of prompting. Classical prompting~\cite{gpt3} has performed poorly on complex reasoning tasks~\cite{wei2023chainofthought}, which is crucial for MTL formalization task.
\begin{figure*}
    \centering
    \includegraphics[width=0.8\textwidth]{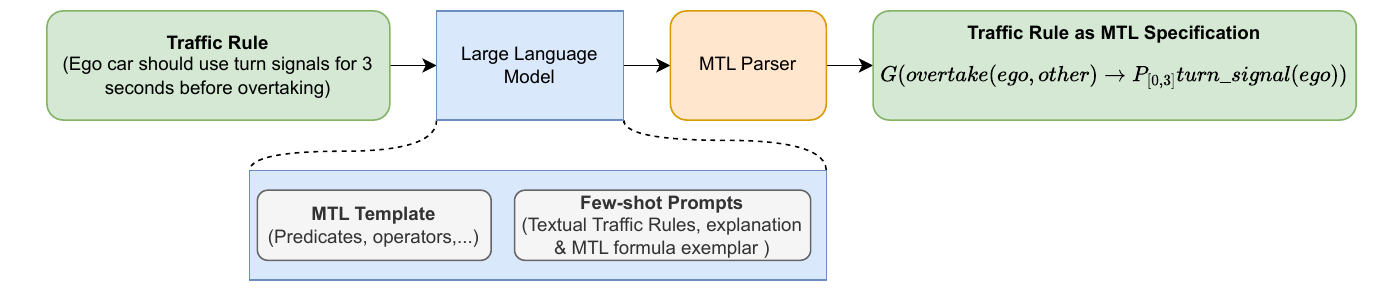}
    \caption{\textbf{TR2MTL architecture for translating traffic rules into MTL.} We use a pre-trained LLM to translate traffic rules into MTL formulas. We pass the traffic rules to the LLM and parse its output with an MTL parser to get the final translation. We guide the LLM with MTL templates and prompts (examples of traffic rules and their MTL translations). Prompts (see Fig.~\ref{fig:noncotprompt},~\ref{fig:cotprompt}) are designed to fit the traffic rule formalization task.}
    \label{fig:architecture}
\end{figure*}

We propose TR2MTL, a framework that translates natural language traffic rules into MTL, as shown in Fig.~\ref{fig:architecture}. The translation module shown here leverages an LLM with a few examples of traffic rules and their corresponding MTL formulas, which are shown as ``few-shot prompts". The LLM also uses an MTL formula template (details in Fig.~\ref{fig:cotprompt}) to guide the translation process. The MTL output of the LLM is then filtered and parsed to generate the final MTL formula. Few-shot prompts shown here use a chain-of-thought (CoT) prompting~\cite{wei2023chainofthought} variant, which divides the translation problem into subtasks, each requiring the LLM to generate an output that solves the current subtask and provides the context for the next one. Such a variant of prompting enables the model to have a more robust reasoning capability. This framework can serve as a valuable assistant, significantly reducing the effort required for rule formalization. It also offers the advantage of enabling users to select the appropriate answer or edit the automatically generated translation from the model rather than starting from scratch to create a new logical rule.

We evaluate TR2MTL using various state-of-the-art and open-source LLMs as the backbone model on a challenging evaluation dataset of traffic rules from StVO and VCoRT. We achieved promising translation results even with relatively complex traffic rules having ambiguous and unstructured syntax and wordings. We also demonstrate the applicability of TR2MTL to various forms of temporal logic and other intelligent vehicle domains for generalization purposes. We provide our dataset of traffic rules and MTL pairs as a valuable resource for further research and development in traffic rule formalization and verification for autonomous vehicles. We envision it as a human-in-the-loop framework that facilitates the rapid formalization of rules and specifications for verification and testing purposes. We can summarize our main contribution as follows:
\begin{itemize}
\item We propose TR2MTL, a novel human-in-loop framework that leverages LLMs and an in-context learning approach to automate the translation of natural language traffic rules into MTL formulas without the need for large annotated datasets.
\item We present a dataset for evaluating the quality of LLM-generated MTL traffic rules. Our dataset consists of unstructured traffic rules, which have varying linguistic complexity and ambiguity, in contrast to curated datasets~\cite{dronePlan}, which are based on pre-processed rules with clear semantics and belong to the non-AV domain. 
\item We demonstrate that our framework can easily adapt to new traffic rules, different logic formalization, and other domains with minimal data requirements.
\end{itemize} 
Sec.~\ref{releated_work} and~\ref{methodology} cover related work and theoretical backgrounds, respectively. Backbone language model choices and dataset are discussed in Sec.~\ref{eval}, followed by the results and discussion in Sec.~\ref{results_and_analysis}.
\section{Related Work}
\label{releated_work}
Selected traffic rules from VCoRT and StVO are manually formalized for intersection~\cite{intersection}, interstate~\cite{interstate}, safe distance~\cite{safedist}, uncontrolled intersections~\cite{karimi_formalizing}, and overtaking situation~\cite{Rizaldiovertake}. Motion planning algorithms also leverage traffic rules, usually in an implicit manner, for rule compliance.

There are multiple ways to formalize traffic rules. Traffic rules are formalized in first-order logic~\cite{karimi_formalizing}. Zhao \emph{et al.}~\cite{onto} used ontology, and~\cite{PL} used propositional logic as a representation choice for traffic rules. However, these representations do not directly use temporal aspects, but workarounds such as implicitly hiding time aspects in the predicate or atomic propositions are used. MTL~\cite{intersection},~\cite{interstate} for a discrete time interval, signal temporal logic (STL)~\cite{stl1},~\cite{stl2} for a continuous time interval of temporal operators, linear temporal logic~\cite{LTL} and Co-safe linear temporal logic (scLTL)~\cite{cosafeltl} are different choices among the temporal logic. MTL is more appropriate for us, as it provides native support for temporal information and is decidable for finite trace length~\cite{alur_logics_1992}. Among these options, LTL and scLTL are unsuitable for modeling the temporal operator's duration. MTL and STL are good choices for modeling rules with temporal aspects, but here we focused on the MTL. However, our approach can be easily extended for STL. In this work, we kept focus on MTL due to our end goal of assisting the trajectory verification system in consideration, but our approach can be easily extended to STL. Furthermore, MTL is more suitable for verifying discrete signals, unlike STL, which works better with continuous signals~\cite{intersection,interstate}.

Formalized rules can be utilized in trajectory monitoring~\cite{intersection},~\cite{interstate}, trajectory planning~\cite{planuse},~\cite{planuse2}, and calculating reachable sets of autonomous vehicles based on rules. We develop an automated framework to formalize traffic rules, which can benefit various use cases such as trajectory monitoring, planning, and reachability analysis for AVs. This differs from previous work that relied on manually defined criteria, which were not flexible enough.

LLMs have become a powerful tool for translation, mathematical problem-solving, and reasoning tasks and are used for reasoning and planning for autonomous agents~\cite{llmasplanner},~\cite{llmrobot},~\cite{manas_srl}. \textit{nl2ltl}~\cite{nl2ltlibm} is developed to translate text to the LTL formula based on the restricted template of the formula with limited capabilities.~\cite{hahn2022formal} fine-tuned the LLM for text-to-LTL and FOL conversion with their dataset, whereas~\cite{stlfinetune} trained model for STL formula generation from textual specification and \textit{lang2ltl}~\cite{lang2ltl} used the in-context learning to obtain the LTL specification for robotics but their methods lack the step-by-step reasoning required for complex rules such as in traffic rules. In contrast, our approach uses in-context learning with CoT prompting instead of classical prompting. This enables us to introduce in-depth reasoning capability for automated MTL formalization of traffic rules. Additionally, the previous work did not work with legal traffic rules as specifications, where texts are unstructured, a lot of world knowledge is assumed, and fine-tuning the model or training from scratch is difficult without sufficient datasets. Previous works have a major focus on LTL, and first-order logic~\cite{hahn2022formal}~\cite{llmasplanner}, having no time constraints over temporal operators or heavily processed.
\section{Method}
\label{methodology}
\subsection{Metric Temporal Logic Representation}
\label{MTL}
We briefly overview the operators in MTL following Thati and Roşu~\cite{thati_monitoring_2005}. The MTL grammar we use in this paper can express MTL specifications as:
\begin{equation*}
\varphi ::= p\mid \neg p \mid \varphi_1 \wedge \varphi_2 \mid \varphi_1 \vee \varphi_2
\end{equation*}
Where $p \in P$ is a set of possible atomic propositions; $\varphi$ is the task specification, and $\varphi_1$ and $\varphi_2$ are MTL formulae. We can have multiple formulas ( $\varphi_1$.....$\varphi_n$).
We also use the following temporal operators that involve time intervals.
\begin{equation*}
\varphi ::= \text{G}_t(\varphi) \mid \varphi_1 \text{U}_t\varphi_2 \mid \text{X}_t(\varphi) \mid \text{F}_t(\varphi) \mid \text{P}_t(\varphi)
\end{equation*}
where $G$, $U$, $X$, $F$, and $P$ are temporal operators, and the subscript $t$ denotes an interval $[\text{t}_1, \text{t}_2]$ representing time duration when these operators are active. If an interval for an operator is not provided, it refers to the end of the trace, which we presume is finite. Formulas can also be written using the logical connectives \textit{negation} ($\neg$), \textit{and} ($\wedge$), \textit{or} ($\vee$), and \textit{implication} ($\Rightarrow$). The future globally operator $G$ ensures that $\varphi$ exists within a time period for all future states. $U$ indicates that $\varphi_1$ holds until $\varphi_2$ becomes true inside the time interval $t$. Operator $X$ indicates that $\varphi$ is valid for the next state within the time span defined by $t$ and $F$ indicates that $\varphi$ holds within a time interval for some future state, whereas the past operator, $P$, specifies that $\varphi$ holds within a time interval for some past or previous state.
\subsection{LLM and Chain-of-Thought Reasoning }
\label{prompt}
\begin{figure*}
    \centering
    \includegraphics[width=0.80\textwidth]{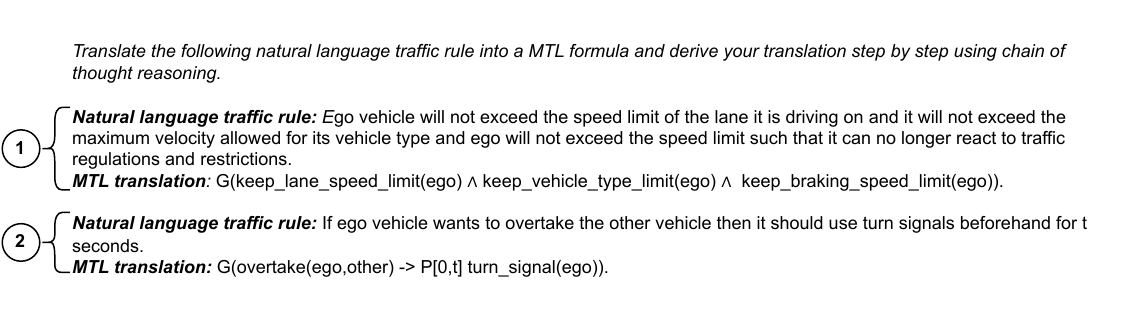}
    \caption{\textbf{Prompt without reasoning and step-by-step thinking for in-context learning.} Two-shot learning prompts or two traffic rules examples are shown. MTL output is directly provided without additional hints. Only Natural language text is provided during the test phase, and the model should generate the MTL formula. Annotations are for illustration (not part of the actual prompts).}
    \label{fig:noncotprompt} \vspace{-5pt}
\end{figure*}

LLM is a statistical model that can comprehend context, infer meaning, and generate coherent responses for diverse sets of tasks. We restrict ourselves to the transformer-based architecture of LLM due to their state-of-the-art performance~\cite{gpt3} in NLP tasks. The transformer architecture uses attention mechanism~\cite{attention} to capture the relationship among word corpus and tokens (words or sub-parts of words) in text, allowing LLM to capture a given word's context and semantics. LLMs were trained on a large corpus of data available on the internet, so they have ingested world knowledge. ``In-context learning or prompting" is a method where the model learns from examples or prompts of a specific task. Prompts consist of task instruction and input-output pairs, as an example. This type of learning is also called \textit{few-shot learning}. If there is only task instruction without any examples, then it \textit{zero-shot learning}. In this work, we used \textit{two-shot learning}, which means we have two examples apart from the instruction provided to the language model. Fig.~\ref{fig:noncotprompt} shows our classical prompt for two-shot learning. We begin Fig.~\ref{fig:noncotprompt} with instructions to the model followed by the two examples of the natural language traffic rule and its MTL formula. The model can then translate new traffic rules based on the learning from these two prompts at inference time, even if new rules and operators are not in the prompt example. This is because LLMs have obtained some knowledge about logic during pre-training~\cite{gpt3} along with knowledge of natural language utterance.

\begin{figure*}
    \centering
    \includegraphics[width=1.0\textwidth]{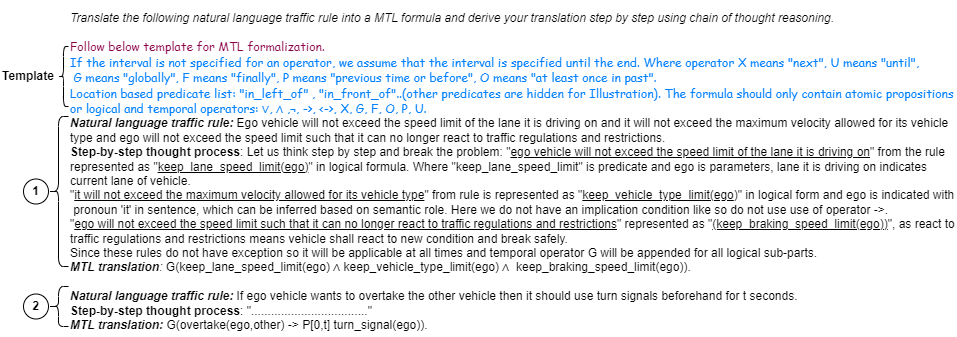}
    \caption{\textbf{Prompt with reasoning and step-by-step thinking for in-context learning.} Prompts used in our experiments, each having two input-thought process-output triplets or two-shot learning (a total of 2 such diverse example pairs were used). The step-by-step thought process breaks the rules into subrules and generates formulas. Later, they are combined to form the complete MTL rules. \textit{Prompts are made of such examples, but only textual rules are provided during testing, and the model generates step thinking and the final MTL formula similar to the two prompts example}. We omit the details of the thought process for the second prompt, but it can be written similarly to the first prompt.}
    \label{fig:cotprompt}\vspace{-10pt}
\end{figure*}

Unlike fine-tuning the LLM, which can be slow and costly, creating the prompts described in~\cite{gpt3} and illustrated in Fig. \ref{fig:noncotprompt} for in-context learning is a good alternative. However, their performance is not optimal for the task that requires reasoning. Inspired by~\cite{wei2023chainofthought}, we introduce CoT prompting with step-by-step thinking, where apart from input-output pairs as in classical prompting, the thought process is introduced, and the new prompt is the triplet of \{input, thought-process, output\}. However, our methods differ from the work of~\cite{wei2023chainofthought} in introducing step-by-step thinking suited for the traffic rule domain in our few-shot prompt creation. In their foundational work~\cite{wei2023chainofthought} and later work focused more on the mathematical reasoning problem, and their CoT reflected the mathematical reasoning. Unlike mathematical language, formal logic involves more open-ended problems that need to follow the syntax and grammar of natural language. Therefore, creating formal language requires mathematical reasoning and linguistic grammar knowledge. Moreover, formal language has more explicit and strict syntax rules. By applying a step-by-step approach to formal language in the CoT process, we can break down a complex traffic rule problem into multiple subtasks of temporal logic creation. Considering the traffic rules, which sometimes span more than one sentence and sometimes do not have a clear structure, breaking the problem into sub-parts or subtasks through step-by-step thinking introduces more reasoning capacity to the model. Our new few-shot prompts have restrictions imposed over the selection of temporal and logical operators through the \textit{template} definition (Fig.~\ref{fig:cotprompt}), which aligns the pre-trained LLMs based on our specific formal logic language requirement. More about the improvement introduced by our variant of CoT can be found in the results section \ref{results_and_analysis}.

The CoT and step-by-step thinking few shot examples used in our in-context learning are shown in Fig.~\ref{fig:cotprompt}. For better readability, we used annotations and highlighted different components of our approach. As shown in Fig.~\ref{fig:cotprompt}, first, instruction is provided to the model to translate text to MTL, followed by the template of our logical language choice (MTL in this case). This template also limits our logic formula based on the logical structure provided to the model. Then, finally, two triplets (more can be added) of \{input, step-by-step thought process, output\} are provided as few-shot prompts. Compared to prompts without chain-of-thought in Fig.~\ref{fig:noncotprompt}, we have additional \textit{step-by-step thought process} in Fig.~\ref{fig:cotprompt}. Visualizing it reveals that in the thought process, we introduce breaking problems into subtasks and solving the subtask; then, we combine them to derive the output for the complete task. It is like solving a mathematical reasoning problem, where breaking a complex problem into subparts makes things easier. In this paper, CoT refers to the chain-of-thought with step-by-step thinking unless otherwise specified. In summary, our model learns to generate MTL translation based on the input rule by decomposing problems and deriving the final MTL translation. This reasoning capability helps the model to perform well with the varied sentence and rule structures. Here, reasoning refers to reasoning over the text and understanding implicit and explicit traffic regulations behind specific rules. With the world knowledge of LLMs, we can make additional assumptions to understand rules and their dependency on any other implicit knowledge. 
\section{Evaluation}
\label{eval}
We evaluate our framework on the dataset created for the domain of the traffic rule. Two kinds of in-context learning prompts- classical prompting without CoT and prompting with CoT—were used in our experiments across several LLM models.
\subsection{Dataset Structure: Textual Traffic Rule and MTL}
\label{dataset}
We aim to formalize traffic rules from unstructured text using MTL. However, we face a challenge in finding a suitable dataset to evaluate our approach. For two reasons, existing datasets and evaluation metrics and evaluation metrics~\cite{llmasplanner}-\cite{hahn2022formal} are not adequate for our purposes. First, to our knowledge, no legal traffic rules have been incorporated as a reference for training or testing their models. Second, they focus on LTL data, which is different from MTL. There are well-curated datasets for robotics and drone motion planning~\cite{dronePlan} based on LTL data. Moreover, existing datasets (even non-traffic rule datasets) have simplified and structured rules that do not require complex reasoning or inference as in traffic rule formalization.~\cite{intersection} and~\cite{interstate} offered a comprehensive codification of MTL traffic laws for intersection and interstate driving. These works are the most relevant to our requirements, and we use them as a baseline for our dataset creation. Their MTL rules are not directly applicable to us, as they are derived from multiple traffic rules of a particular subtype and depend on the legal analysis of various rule sources (StVO, VCoRT) and court decisions for specific traffic scenarios. This rule representation is helpful for the manual approach as their approach tends to capture as many traffic situations in less number of rules. However, they are unsuitable for LLM-based unstructured traffic rules to MTL translation. Datasets created for other domains (robotics and drones) are good examples of rules that can be adapted and used for manual and automated models. In addition, these MTL rules are incompatible with learning-based models such as LLMs, and methods are hard to extend to new rule sets. So, we created a new dataset of traffic rules and MTL pairs that can work with LLMs. We kept the original format and language of the legal traffic rules text with minimal preprocessing or simplification. We also incorporate a few important traffic rules from earlier works on AV planning and forecasting~\cite{planuse2},\cite{ruleguidetp}. However, the dataset does not cover some less representative or exceptional driving scenarios in great detail.

A dataset of 50 traffic rules and MTL pairs was generated based on the methodology explained in~\cite{manas_iclp} and MTL syntax described in Sec.~\ref{MTL}. As in~\cite{manas_iclp}, we leveraged the parsed tree structure to develop the MTL formula, allowing us to find reusable predicates that can be used for multiple formulas. Our dataset handles both general driving rules and rules with exceptions where the current rules are not valid due to new situations on the road, such as emergency vehicles' right of way. Legal text traffic rules were derived from VCoRT (for general driving rules) and the StVO (for country-specific rules, Germany in this case). Some preprocessing is given to the rules to give the phrases concrete meaning and remove the rules' ambiguity. As per StVO §(3)(1), a vehicle may only drive at a speed that allows for constant control. In formal logic and automated systems, ambiguous phrases like \textit{under control} require definition. We preprocess such phrases with more concrete terms, such as speed less than $max\_speed\_limit$, to avoid multiple user-dependent interpretations.
\begin{table*}[h]
    \caption{OVERVIEW OF TRAFFIC RULES AND MTL PAIRS DATASET}
    \label{tab:traffic_rules}
    \begin{tabular}{ p{.48\textwidth} p{.44\textwidth}  }
        \toprule
\textbf{Natural Language Traffic Rule }     
& \textbf{MTL Representation} \\\midrule
The ego vehicle passes a stop line if the stop line is in front of the ego vehicle and is not in front of it at the next time step.~\cite{intersection}        
& $\begin{aligned} &{cross(ego,stop\_line)} \\ &\quad\Rightarrow ({in\_front(stop\_line,ego)} \land X(\lnot{in\_front(stop\_line,ego)})) \end{aligned}$\\\hline
At roundabouts and junction, if there is a pedestrian crossing the street or a cyclist on the left of the ego vehicle, then the ego vehicle must give way to the pedestrian or cyclist, unless there is a traffic sign number 205 indicating otherwise. & $\begin{aligned}
& G((at\_roundabout(ego) \lor at\_crossroad(ego)) \\
& \land (at\_pedestrain\_crossing(pedestrian) \lor left\_of(bicycle,ego)) \\
& \land \neg at\_traffic\_sign(ego,205) \\
& \quad\Rightarrow yield(ego,(pedestrian\lor bicycle)))
\end{aligned}$\\\hline
Ego vehicle will not exceed the speed limit of the lane it is driving on and will not exceed the maximum velocity allowed for its vehicle type and ego will not exceed the speed limit such that it can no longer react to traffic regulations and restrictions.
& $\begin{aligned}
&G({keep\_lane\_speed\_limit(ego)}\\ &\quad\land{keep\_vehicle\_type\_limit(ego)}\\
&\quad\land{keep\_braking\_speed\_limit(ego)})
\end{aligned} $ \\
        \bottomrule
    \end{tabular}\vspace{-7pt}
\end{table*}
Some rules refer to other rules in the legal text, so we preprocessed them to make them clearer and more complete by merging the fragments across the legal text. We minimally preprocessed the rules to preserve their complexity so that the model can handle unstructured natural language rules. We paid special attention to capturing the complex linguistic semantics of the unstructured natural language text. This enabled us to cover rules with unclear implications, rules with conditions that were fulfilled in the past and have ramifications in the present, long or multiple sentences with complex pronouns and verb structures, and rules where it was not clear whether the conditions were connected by logical ``OR" or logical ``AND".

Our datasets also uses predicates and arguments from commonroad~\cite{commonroad} and lanelet~\cite{lanelet}. We assume the sensor detects objects correctly, and a high-definition map provides information about road infrastructure. TR2MTL can be adapted to other dataset structures by providing a list of available predicates and arguments based on the information of the other datasets. Our goal is to provide a framework and a dataset for automatically translating traffic rules into MTL through LLM and to evaluate how well TR2MTL can encode both temporal operators and their timing constraints.

Table~\ref{tab:traffic_rules} shows our dataset's natural language traffic rules and their corresponding MTL. As evident from the data sample in Table~\ref{tab:traffic_rules}, traffic rules exhibit a distinct linguistic structure that sets them apart from other temporal logic datasets. These rules tend to be longer and more complex. For instance, consider sample rule 2 in Table~\ref{tab:traffic_rules}: ``At roundabouts and crossroads, if there is a pedestrian crossing the street or a cyclist on the left of the ego vehicle, then the ego vehicle must give way to the pedestrian or cyclist, unless there is a traffic sign number 205 indicating otherwise." Upon closer examination, we observe that despite the mention of both roundabouts and crossroads in the rule, from a logical perspective, it implies a ``logical OR" condition. It means that even if only one of the conditions (roundabouts or crossroads) is detected through driving sensors or map data, the rule should still be applicable. This shows the complexity that TR2MTL will face while formalizing traffic rules. Our formalized dataset and steps to reproduce traffic rules and MTL pairs can be accessed\footnote{Evaluation dataset and model prompts can be accessed here:\url{https://github.com/kumarmanas/traffic_law_formalization}}. We will continue adding new traffic rules to support the growing community of researchers working at the intersection of NLP, LLMs, and rule compliance for developing safe and intelligent AVs. We acknowledge that in its current form, our dataset is more suitable for evaluation rather than for fine-tuning the model due to its small size. Nevertheless, it is a good resource for in-context learning models and trained model evaluation for AV and MTL domain.
\subsection{Evaluation Setup}
\label{eval_setup}
In our work, we performed two sets of in-context learning experiments: one involving CoT prompting and the other without such prompting. However, unless otherwise specified, we discuss experiments involving CoT prompting due to the better result. As shown in Fig. \ref{fig:architecture}, we parse the language model-generated output with MTL parser py-metric-temporal-logic\footnote{\url{https://github.com/mvcisback/py-metric-temporal-logic}}. This post-process allows us to generate a better candidate MTL formula. Additionally, the template provided for the prompts (see Fig. \ref{fig:cotprompt}) restricts the grammar of the generated MTL formula. We can swap out the MTL parser and filter module for a new logic parser module, say for parser of LTL or STL language.
\subsection{Model Selection}
\label{model_selection}
We introduce our selected LLMs, which are used as a backbone for this work. As explained in the introduction, we selected the ``in-context learning" approach. In-context learning via prompting, as introduced in Sec. \ref{prompt}, is an alternate way to use the language model to learn new tasks given only a few examples. We give LLM a \textit{prompt} consisting of a list of input-output pairs that demonstrate the task. Five different language models are used as LLM backbone in TR2MTL: \textbf{\textit{GPT-4}\footnote{\label{openaimodel} gpt-4-0314 and gpt-3.5-turbo-0613  models are used for GPT-4 and GPT-3.5-turbo, respectively.}, \textit{GPT-3.5-turbo}\textsuperscript{\ref{openaimodel}}, \textit{StarCoder}~\cite{Li2023StarCoderMT}, \textit{Falcon-7b}~\cite{falcon} and \textit{Bloomz}~\cite{bloomz}}. We selected models based on three criteria: state-of-the-art performance in multiple domains, task relevance, model accessibility, and open-source availability. Not all models met all these criteria. GPT-4 and GPT-3.5-turbo were accessible state-of-the-art models when writing this paper but not open-source. On the other hand, StarCoder, Falcon-7b, and Bloomz models are open-source with publicly accessible API. Bloomz and Falcon-7b are optimized to follow human instructions for various tasks. Whereas the StarCoder performs very well for the coding tasks, even with smaller model sizes~\cite{Li2023StarCoderMT}. We evaluated the code generation model to evaluate if it can be extended to formal logic.
\subsubsection{Reducing inconsistencies in TR2MTL Output}
LLMs are statistical models that may generate inconsistent output for the same task. We can not eliminate this problem, so to limit its impact, we used a temperature value of $0.25$ and $P=1$ so that in multiple runs, it generates more consistent output. Temperature can range from $0$ to $2$, and lower values produce more consistent and deterministic output (needed for safety systems) but less creativity. We chose $0.25$ as it gave us the best result. Templates (shown in Fig.~\ref{fig:cotprompt}) and a list of allowed predicates also helped us reduce inconsistencies, but not entirely. For example, the MTL formula for a rule about overtaking had inconsistent predicate generation without the template in the prompt. The ideal MTL formula for one of the overtaking rules is $G(overtake(ego,other) \rightarrow P_{[0,t]}turn\_signal(ego))$, but TR2MTL sometimes generated \textit{overtake} and sometimes \textit{overtaking}. We reduced this kind of problem by providing a list of predicates.
\section{Results and Analysis}
\label{results_and_analysis}
Table~\ref{table:eval_table} shows that our framework performs better on our traffic rules dataset by using CoT, which leverages contextualized temporal operators. We compare different LLM backbones and find that GPT-3.5-turbo with CoT has the highest accuracy. Remarkably, this result is obtained with only two prompts, demonstrating the effectiveness of our approach, and more prompts could lead to even higher accuracy~\cite{wei2023chainofthought}. We use nl2ltl as a baseline, which converts natural language to LTL formulas without time constraints. We simplified the MTL formula to an LTL formula for this evaluation by omitting the exact time information of the temporal operators. However, nl2ltl does not support CoT, which is essential for our framework. We consider a translation incorrect if it has any of these errors: missing or wrong temporal operator, wrong predicate, wrong argument order, wrong logical connective, or MTL grammar violation. We consider a translation correct if it uses the correct predicates and has the same meaning as the ground truth translation, even if the MTL formalization is different. We treat $yield(ego,other) \vee right\_of(other,ego)$ and $right\_of(other,ego) \vee yield(ego,other)$ as the same translation since they have the same logic. Likewise, $right\_of(ego,other)$ and $left\_of(other,ego)$ are also equivalent. To scale the accuracy calculation for larger datasets, we can use either a model checking tool or a semantic accuracy metric as in~\cite{hahn2022formal}. To ensure the standardization of our evaluation, we implemented a robust validation process. This involved running the TR2MTL model five times for each input and selecting the translation that was chosen by majority vote. This approach helped to mitigate any potential inconsistencies in the model's output. Additionally, we manually inspected the translations to verify their accuracy and extended them to ensure that the generated formulas were parsable by MTL parsers.
\begin{table}[htbp]
\begin{center}
\caption{\textbf{Translation Accuracy of models for MTL translation of traffic rules.} Model is evaluated for two types of in-context learning: with and without CoT. Bold values represent the top models for each experiment.}
\label{table:eval_table}

\begin{tabular}{lcc}
\toprule
\textbf{Model Architecture} & \textbf{With CoT} & \textbf{Without CoT}\\
\midrule
TR2MTL (with GPT-4\textsuperscript{\ref{openaimodel}}) & \textbf{72.91}\% & \textbf{43.75}\%\\
TR2MTL with GPT-3.5-turbo\textsuperscript{\ref{openaimodel}}) & 66.66\% & 39.58\%\\
TR2MTL (with StarCoder~\cite{Li2023StarCoderMT}) & 50.00\% & 12.50\%\\
TR2MTL (with Falcon-7b~\cite{falcon}) & 37.50\% & 12.50\%\\
TR2MTL (with Bloomz~\cite{bloomz}) & 22.91\% & 10.41\%\\
nl2ltl\cite{nl2ltlibm} & - & 29.16\%\\
\bottomrule
\end{tabular} 
\end{center}
\end{table}\vspace{-5pt}
Our dataset is designed to be challenging using longer sentence instances and semantic complexity as explained in \ref{dataset}. Analyzing incorrect translations among all models revealed that most have longer sentence instances, sometimes exceeding language model capability. This failure is more apparent in non-GPT model variants, where longer sentence translation either provides erroneous translation or the model can not produce translation at all. In traffic rules such as \textit{At roundabouts and crossroads, if there is a pedestrian crossing the street or a cyclist on the left of the ego vehicle, then the ego vehicle must give way to the pedestrian or cyclist, unless there is a traffic sign number 205 indicating otherwise}, we also have a complex semantic structure. There is much logic to unpack for the model. GPT-4 and GPT-3 based models generate better quality translations compared to the other models. One possible reason behind this good performance is the vast background world knowledge available to the model. For this traffic rule, only GPT variants of the model can infer that we need logical OR instead of logical AND for the part of rule ``at roundabouts and crossroads...", as based on general knowledge, we tend to know such distinction. We also observed that the code generation model, StarCoder, did not perform well where additional background information was required. To express traffic rules in formal logic from natural language, one needs to know about the world and driving knowledge, reasoning logically, and using world knowledge to support the implicit rules. So, state-of-the-art non-code LLMs are better equipped to handle than code generation models in the absence of fine-tuning data.

\begin{figure}
    \centering
    \includegraphics[width=0.3\textwidth]{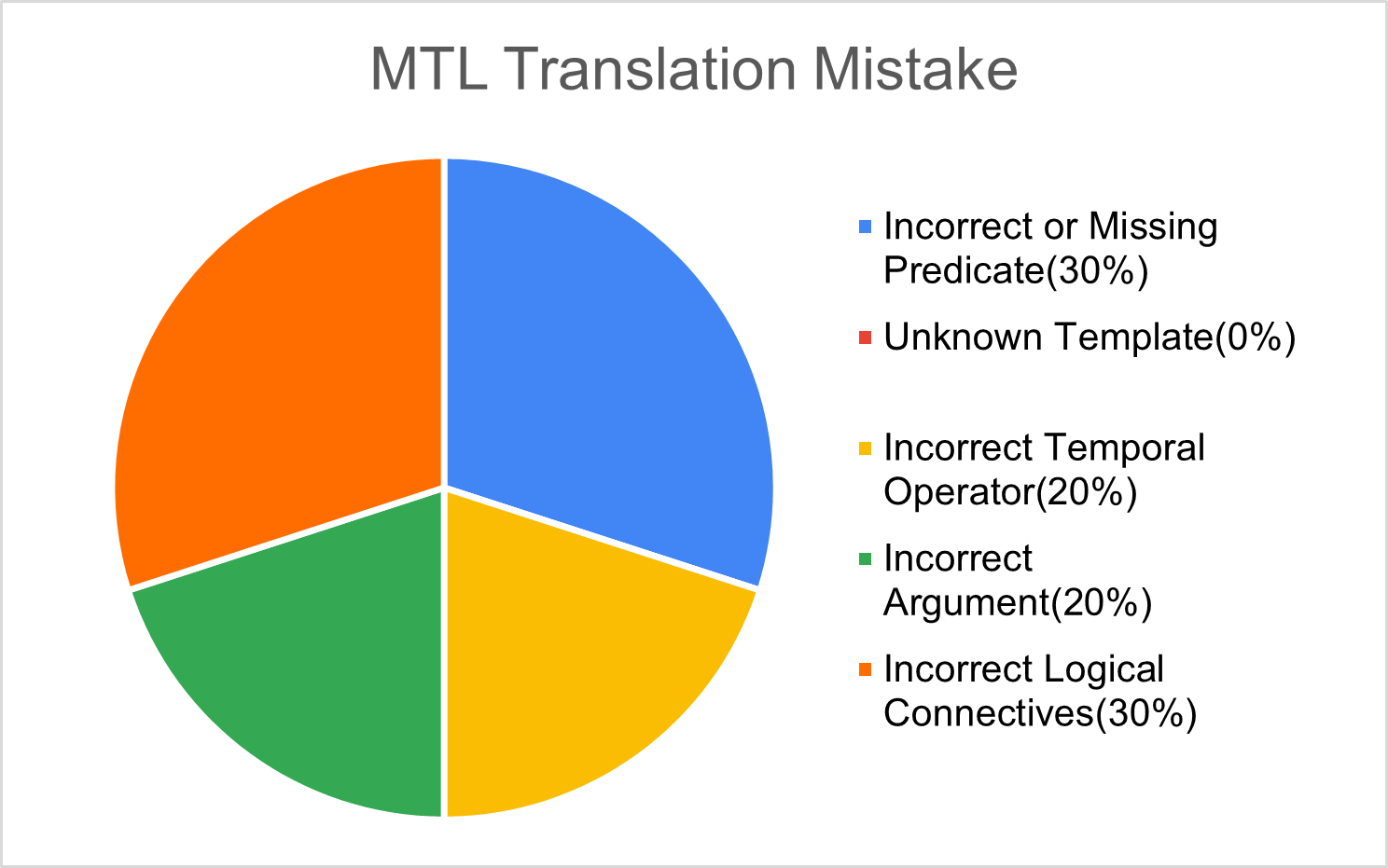}
    \caption{\textbf{Distribution of mistakes made by the framework.}}
    \label{fig:mistake}\vspace{-15pt}
\end{figure}

We performed an error analysis of MTL formalization, and Fig.~\ref{fig:mistake} shows the error distribution with respective error types for our best-performing GPT-4 model with the CoT reasoning method. All translations were per MTL grammar, indicated by $0$\% mistake for ``unknown-template". This can be attributed to the template provided in the prompts and parser module. After analyzing all the model's results, we found that most errors belong to incorrect logical connectives (including incorrect implication operator) followed by incorrect predicate. An instance of an incorrect translation is when the allowed predicate is fused with its argument. For instance, instead of $in\_front(ego, stop)$ TR2MTL represents it as $stop\_in\_front(ego)$. In addition, our model provides step-by-step output with a dictionary having a collection of traffic rule components (usually as smaller fragments of text rule) and their corresponding propositions. This makes the result more explainable and easy for the end user to use or edit as editing is easier for the end user than creating a new rule.

CoT is restricted by the number of tokens or length of prompts that the model can process, which might hinder the user's ability to write long or more prompt examples. This prevents us from adding more context to CoT. We can improve the accuracy of TR2MTL by using a semantic parser as a pre-processing module to transform traffic rules into a more controlled language (e.g., logical English) format needed for formal logic.
\subsection{Generalization Capability}
\label{gen}
We used a few shot prompts with just two examples, which poses the question of the generalization capability of TR2MTL for different logic structures, temporal operators, and out-of-distribution tasks. Table~\ref{table:eval_table} indicates that our framework can generalize it to unseen MTL operators (e.g., X, F..) in the prompts but provided in templates. It can also handle sentence structure and complexity not seen in two prompts. It can also generalize to new entities. For example, we use ego and other as two types of arguments in our prompts. However, the model can differentiate between vehicles in front of the ego and those behind the ego. However, we observed that code-generation models without CoT are more susceptible to memorization problems than noncode-generation models. We tested our framework on six \textbf{marine traffic rules} from~\cite{marine} and achieved $50$\% accuracy. This shows that our framework can handle different types of rules for the domain of other intelligent vehicles. 

To demonstrate our approach's robustness on a larger sample, we evaluated the drone planning dataset~\cite{dronePlan}, which contains 810 pairs of natural language instructions and LTL formulas. Our goal was to demonstrate the effectiveness and extensibility of our approach to other logics (such as LTL) and domains. Table~\ref{tab:drone} shows that our CoT method achieved reasonably good accuracy with just 7 example pairs, compared to the state-of-the-art fine-tuned model~\cite{dronetestfine}. The fine-tuned model typically has higher accuracy, as it benefits from more data than in-context learning. This indicates the potential of our approach for other domains that rely on rules to perform their tasks.

\begin{table}[h]
\begin{center}
\vspace{-8pt}
\caption{Evaluation of the accuracy of translating planning instructions to LTL for the drone planning dataset. \textit{Blue color indicates prompting-based model and brown indicates fine-tuned model.}}
\label{tab:drone}
\begin{tabular}{lccc}
\toprule
\textbf{Model backbone} & \textbf{Acccuracy}\\
\midrule
\textcolor{blue}{TR2MTL (with GPT-4\textsuperscript{\ref{openaimodel}})}  & 57.91\%\\
\textcolor{blue}{TR2MTL (with GPT-3.5-turbo\textsuperscript{\ref{openaimodel}})} & 51.43\%\\
\textcolor{blue}{TR2MTL (with StarCoder~\cite{Li2023StarCoderMT})} & 42.59\%\\
\textcolor{brown}{BART-FT-RAW~\cite{dronetestfine}}  & 69.36\%\\
\bottomrule
\vspace{-25pt}
\end{tabular}
\end{center}
\end{table}
\subsection{Prompt Selection and Translation Accuracy}
Prompt selection is crucial for translation in the low-data domain. Prompts should cover a diverse range of traffic rules and MTL structures rather than linguistic variations, as LLMs are more proficient in natural language than formal logic. We observed that TR2MTL struggles with MTL formulas that have different logical structures from the prompts. For example, if the prompts have the structure $G(A \lor B) \lor C$, TR2MTL performs poorly on formulas with the structure $G(A \vee B) \land C$. We also need to consider the diversity of arity in the prompts, as TR2MTL fails to generate formulas with more than one argument for a predicate if the prompts only have one argument. The reason is that we have limited data for prompting, so we need to sample prompts based on the final task and the LLM’s capabilities. We can create prompts that target the LLM’s weaknesses and improve its performance.
\subsection{Trajectory Monitoring Using MTL Rules}
\label{monitor}
We used formalized rules for trajectory monitoring. Our framework can be extended to STL, which also uses timing constraints over temporal operators. To demonstrate our approach, we generated rule-compliant and non-compliant trajectories using CommonRoad. We verified them against the specifications using a runtime Python monitor and an MTL parser as in~\cite{marine} and~\cite{manas_iclp}. The only modification required from TR2MTL is to write some additional functions to map the high-level predicates to the CommonRoad format for the final trajectory evaluation. Future work can focus on automating this mapping process.
\section{Conclusion and Future Work}
An automated human-in-loop framework for legal traffic rule formalization using a language model is presented. Due to our CoT prompting, our approach can quickly adapt to new rulebooks in the AV domain or other intelligent vehicle specification formalization tasks. Formal specifications are essential for AVs to comply with legal requirements, and our approach simplifies this process and makes output more understandable. Our method was applied to traffic rules requiring reasoning over text, and we observed that the language model could leverage its knowledge to produce safe and lawful specifications to a certain extent. Still, further enhancement is needed as verification is a safety-critical domain. Currently, our framework is more of a human-in-loop system. Furthermore, our method can create different types of formal rules (STL, LTL, etc.) by changing the prompts and parsers, and it can aid traffic management and manual coding tasks. We can also improve the performance of our method by adding more context to CoT~\cite{contextwindow} or more prompts~\cite{wei2023chainofthought}. In future work, we plan to apply and ground the generated formulas to the sensor data of AVs and evaluate end-to-end rule-to-trajectory verification. We aim to add more traffic rules to the dataset to fine-tune the language model and improve its performance.
\bibliographystyle{IEEEtran}
\bibliography{ref}

\end{document}